\documentclass[conference]{worldcomp}

\usepackage[hmargin=.75in,vmargin=1in]{geometry}
\usepackage[american]{babel}
\usepackage[T1]{fontenc}
\usepackage{times}
\usepackage{caption}

\usepackage{textcomp}
\usepackage{epsfig,graphicx}
\usepackage{xcolor}
\usepackage{amsfonts,amsmath,amssymb}
\usepackage{fixltx2e} 
\usepackage{booktabs}

\newcommand{\url}[1]{#1}

\title{\bf Pattern Recognition in Collective Cognitive Systems: 
Hybrid Human-Machine Learning (HHML) By Heterogeneous Ensembles}           

\author{
{\bfseries Hesam T. Dashti$^1$, Adel Ardalan$^2$, Alireza F. Siahpirani$^3$, Jernej Tonejc$^1$, Ioan V. Uilecan$^1$,}\\ 
{\bfseries Tiago Simas$^4$, Bruno Miranda$^4$, Rita Ribeiro$^4$,  Liya Wang$^5$, and Amir H. Assadi$^{1*}$}\\
$^1$Department of Mathematics, University of Wisconsin, Madison, WI, USA\\
$^2$Database Research Group, Electrical and Computer Engineering Department, University of Tehran, Iran\\
$^3$ Department of Mathematics, University of Tehran, Iran\\
$^4$ UNINOVA-CA3, Campus FCT/UNL, Portugal\\
$^5$ Cold Spring Harbor Laboratories, NY, USA\\
}

\begin{document}

\maketitle                        

\footnotetext[1]{
Corresponding author: Amir Assadi (\url{ahassadi@wisc.edu}).\\
All co-authors contributed towards the mathematical computations, design of
algorithms, and development of software. This research is partially
supported by NSF-BIO DBI in 2006-07, and NSF-DMS SCREMS since 2009.
The authors are grateful for helpful scientific discussions in plant
biology and genetics by Professor Patrick Masson. Early parts of this
research has also benefited from discussion of data acquisition and sample
root images from gravitropism experiments by Nathan Miller and Tessa
Durham-Brook images from root branching and growth of lateral roots by
Candace Randall Moore. The authors have benefited from generous
collaboration and technical advice by R\&D staff at nVIDIA Inc., SUN
Network.com (currently SUN-Oracle) and the SUN-Grid computation grant from
Sun Microsystems Inc. The authors gratefully acknowledge the generous help
in engineering of the automation hardware and research collaboration by PBC
Linear Inc.}

\begin{abstract}
The ubiquitous role of the cyber-infrastructures, such as the WWW, provides myriad opportunities for machine learning and its broad spectrum of application domains taking advantage of digital communication. Pattern classification and feature extraction are among the first applications of machine learning that have received extensive attention. The most remarkable achievements have addressed data sets of moderate-to-large size. The 'data deluge' in the last decade or two has posed new challenges for AI researchers to design new, effective and accurate algorithms for similar tasks using ultra-massive data sets and complex (natural or synthetic) dynamical systems. We propose a novel principled approach to feature extraction in hybrid architectures comprised of humans and machines in networked communication, who collaborate to solve a pre-assigned pattern recognition (feature extraction) task. There are two practical considerations addressed below: (1) Human experts, such as plant biologists or astronomers, often use their visual perception and other implicit prior knowledge or expertise without any obvious constraints to search for the significant features, whereas machines are limited to a pre-programmed set of criteria to work with; (2) in a team collaboration of collective problem solving, the human experts have diverse abilities that are complementary, and they learn from each other to succeed in cognitively complex tasks in ways that are still impossible imitate by machines. Thus, from an abstract viewpoint, in solving complex visual perception-cognition problems, a hybrid network of humans and machines could be far more powerful than a network of intelligent machine-only agents whose capabilities are bounded by what the present state of AI knowledge could offer. This article reports a preliminary progress towards theoretical foundations for HHML as a semi-programmable case belonging to the broader domain that we refer to as 'Collective Cognitive Systems'. We use an intuitive composition of ANN building blocks, which is inspired by functional components within a natural neuronal topography, and incorporates pseudo-layers. The performance of this exploratory framework is tested on two datasets, one from biology and one from astronomy with promising outcomes
\end{abstract}



\section{Introduction}
Cybernetics as a recognized field of research emerged in the 1940s-50s as a domain of inquiry. Its raison d'etre at the time was to answer the 'why' and the 'how' of the awesome performance of biological brains, which were metaphorically regarded as performing computations and other logical operations. This era coincides historically with maturity and endowment of abstract mathematics (foundations of mathematics, set theory, logic) and its entanglement with theoretical physics (quantum theory, the nature of space-time-matter) and theoretical biology (life as a physical phenomenon). In the 1960s and 70s, Artificial Intelligence attracted much attention with the ambitious promise of programming computers to perform human-like pattern recognition and other perceptual-cognitive tasks, such as in human vision. Fundamental theoretical obstacles on this extraordinary claim affirm broader superiority of human cognitive performance in 'natural tasks' over similar attempts by computing machinery. Theoretical understanding and rigorous mathematical delineation of the limitations of machine learning are at the very heart of AI, and a prerequisite for discovery of alternative technological solutions to overcome such obstructions. \\
Foundational contributions of Vladimir Vapnik \cite{c1,c2} and others revealed the advantages of inclusion 'empirical' concepts, and opened the way for machine learning to take advantage of the human factor beyond conventional ANN architectures, frequency-based and Bayesian statistics. In a related direction and independently, neuroscientists H. Barlow \cite{c21}-\cite{c23} and others made seminal contributions to the role of sparse coding and information in brain's performance. W. Bialek \cite{c24,c25}, T. Poggio \cite{c26}, T. Sejnowski \cite{c27} and others bridged the gap between the biological-behavioral and the computational-mathematical models of brain function utilizing any or all of the above-mentioned concepts \cite{c3}-\cite{c10}. In the last decade of the 20th century, a different breakthrough in brain research emerged, and to this date, it continues to receive much attention. D. Field, B. Olshausen and others \cite{c11, c12} took a dramatic turn in the experimental approach to study the brain while performing tasks in 'natural scenes'. This was a breakthrough in thinking about intelligence in its natural setting, and outside the stringent conditions of the conventional labs. There are myriad ideas and no shortage of improvements and synthesis of the many fruitful accomplishments of biological and computational learning. The unique medium, however, offered by broadband, ubiquitous digital communication and the ensuing cyberspace requires a fresh approach to two centuries of milestones in studying intelligence and intelligent behavior. 
This article takes the historical viewpoint that each of the above-mentioned conceptual breakthroughs "the biological, physical and mathematical structures" have been developing towards the following common regimes: (a) the nature of intelligence and intelligent behavior are dynamic in the sense of physics; the patterns that are observable in a physical approach to biological intelligence are mathematically associated to Complex Systems, thus continually subject to indeterminacy, reasoning under uncertainty and modeling in a probabilistic framework. (b) Such systems have hierarchical organizations at multiple scales, and observations of different levels of the hierarchy require multi-resolution accuracy. (c) The adjacent levels of hierarchy are brought together by non-linear interactions (i.e., require more than the 'superposition and scaling' that is the hallmarks of any linear theory). (d) Quantifiable communication schemes most probably govern the rules of interactions within and among the hierarchy (sub-) elements. (e) The physical units of quantitative communications for each scale and level of the hierarchy could be 'reconstructed' from sufficient numbers of observations from the smaller-scale dynamics and behaviors. The unit in one level of the hierarchy is often the Gestalt of an ensemble of entities from physically smaller-scale entities of the hierarchy. \\
With these preliminaries out-of-the way, we proceed to computationally explore a realization of the concept appropriate for the current state of machine learning, namely, the intelligent behavior emerging from a collection of intelligent agents that are not constrained to be exclusively biological or machine-like. This is referred to as a heterogeneous ensemble in view of the fundamental differences between the human and the machine intelligence, as outlined above. Moreover, the intelligent behavior is expected to be along the lines of Field-Olshausen approach \cite{c11, c12} to take on 'natural scenes and natural stimuli' to probe the nature of intelligent performance. Accordingly, we study the 'natural pattern recognition' tasks of the type initiated by one or more biological intelligent agents. Indeed, we emphasize the importance of the physical medium for communication as an equally important factor in the exploratory research below. Finally, the physical currency of communication is what we refer to as 'biological information', while the dynamic process of its manipulation and successfully reaching to a stopping criterion (a solution) is referred to as 'biological computation'. The progress report on biological communication and computation are relegated to other forthcoming articles. The reader, however, could trace the germ of history of idea development in the modest preliminary account below. In the next section we illustrate a practical example of the Collective Cognitive System, and the achieved results of this example are described in the "Experimental Results" section.

\section{Methods}
\begin{figure}[htb]\centering
\includegraphics[width=\columnwidth]{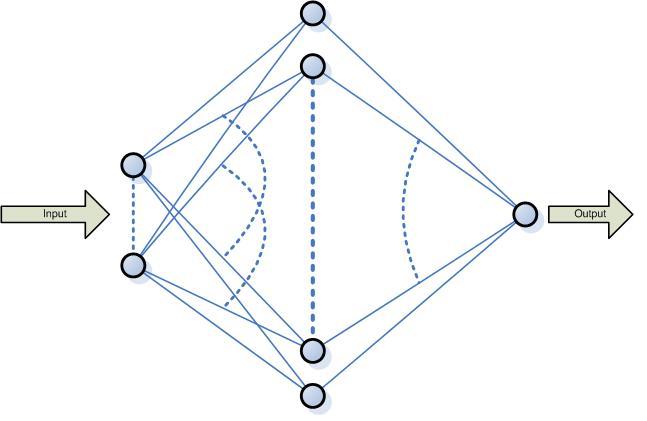}
\caption{\small UBP topology includes three layers: input, hidden, and output. This is the basic back propagation ANN that is used in Hyper-Layers. For simplicity just one hidden layer is shown.} \label{fig1}
\end{figure}

The Artificial Neural Networks (ANN) is inspired by the structure and performance of higher animals brain. The human brain with its sophisticated topology can determine and extract significant features of objects, so it is intuitively reasonable to simulate a similar structure for the feature extraction problem as mentioned in the Introduction section. This HHML methodology is an example of the Collective Cognitive System as described above. \\
The HHML method is composed of a Super Structure Artificial Neural Network (S2AN2), which would be trained like an ANNs for classification \cite{c13}. The training process would be performed on labeled data sets with at least two different classes and the process adjusts weights of the network towards performing the training purpose. As proved in \cite{c14}, weights of a trained ANN represent amount of transitory impact of its corresponding nodes, as the purpose of training process (for a sample classification problem; from now on the purpose of training process is training a classifier). Based on these values, the weights form a ranking of the S2AN2 nodes as well as the nodes in the input layer. Moreover, this ranking can be used to arrange associated features to the input nodes, which represent effectiveness of those features with respect to our training goal. The key idea of the HHML is to study this effectiveness and refine/reduce the set of features. Reexamining the S2AN2 with the reduced set of features and considering precision of its results, gives a good evaluation of the correctness of the method, which is considered in the "Experimental Result" section.\\
The S2AN2 uses the Back Propagation algorithm and is composed of two hyper-layers; the first one is designed to study effectiveness of features. In this layer, for each class there is a Unit Back Propagation ANN (UBP) that gets all the inputs (features) and, by the end of the training process, each of these UBP's shows how much every feature has been decisive for learning its associated class. On the next step, for a training dataset with K classes, the second hyper layer includes a UBP with K inputs and shows object ID's in its output layer. For each UBP, the number of hidden layers and their nodes is in direct relationship with original number of features. The employed activation function is the common sigmoid function. A template of a UBP is depicted in Figure\ref{fig1}.

\begin{figure}[htb]\centering
\includegraphics[width=\columnwidth]{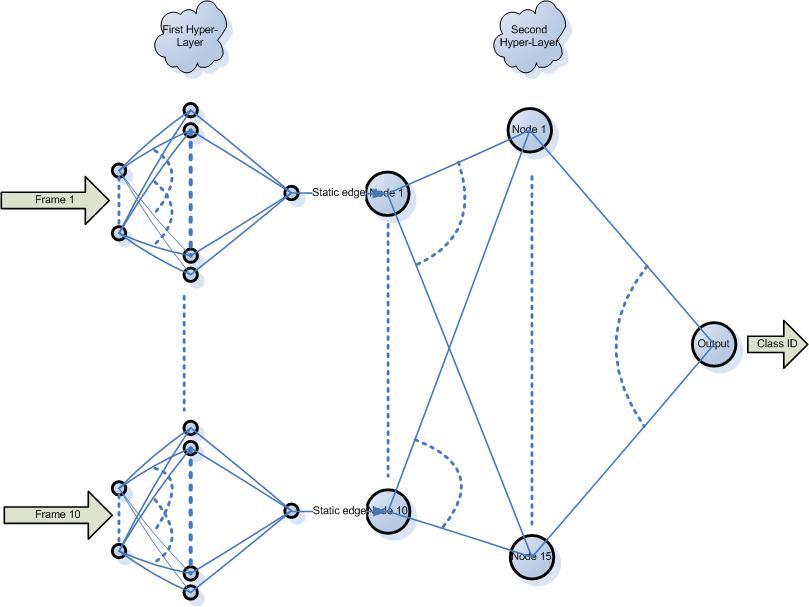}
\caption{\small Topology of a sample S2AN2 realization. For this example, it is assumed that there are 10 classes of objects, so there are 10 UBP in the first hyper-layer. The outputs of the first Hyper-Layer are connected directly to the corresponding input node in the second Hyper-Layer.} \label{fig2}
\end{figure}

As mentioned, there are 'K' UBPs in the first hyper-layer, where all the features fed to all of them. In this layer each UBP has an output node that returns a real number between [-1, 1] corresponding to the amount of collaboration of that UBP for calculating the corresponding class ID of the input. Obviously, one can train the network such that assigns each UBP to a class. The output node of a UBP takes the value '1' when the object belongs to the class associated to that UBP and this value goes to '-1' as far the uncertainty about the class membership goes up. This is a floating point number respective to the similarity of classes. Output nodes in this hyper layer are connected to input nodes of the second hyper layer with constant edge (having the constant weight of 1). These edges are assigned a constant value in order to be neutral in the course of the learning process while connecting the two hyper-layers. During the training process, the UBP in the second hyper-layer analyzes outputs of the K previous UBP's and based on them determines class ID's. The UBP in the second layer calculates its error, update its weights and back propagates errors to the previous Hyper-Layer. Based on these errors each UBP in the first Hyper-Layer calculates its local error and propagates the error for updating its own weights. The topology of the S2AN2 is given in the Figure\ref{fig2}.\\
Our method can be categorized as a zero-order method of model-dependent feature selection, which uses the network parameters only. This means that the selection of the important features of an input is decided by considering only the weights of our specific structure. This gestalt is a batch process in which we design a fully connected network for each class and train them to specifically handle their own class objects. Moreover, results of these networks are processed using another network (in the second hyper-layer) towards determining class ID. The whole structure could be viewed as a homogeneous ensemble of ANNs that use them both as building blocks and the final integrator. This has the very useful advantage of structural and computational homogeneity that makes it suitable for parallel hardware design and implementation, and in turn, yields into super fast feature selection (special purpose) hardware.
\section{Experimental Results}
For a dataset of objects inside of a predetermined feature space, a feature extraction algorithm is supposed to choose a subset of features such that this subset can capture whole (or as much as possible of) objects' information. The HHML algorithm is examined on classification problem of two different data sets: one from astronomy and another from plant biology, where experts labeled them (provided the class ID's) using a priori knowledge. The former comes from light curves of the SMC stars \cite{c15} from the OGLE mission  \cite{c16}. The latter is a data set of root growth movies and consists of two classes of roots; namely wild type and mutant root seedlings. Following the process of the dimensionality reduction via the HHML method is illustrated subject to the data sets. In both cases the precision values are calculated based on the common formula:\\
\begin{center}
Precision = $\frac{True Positive}{True Positive + False Positive}$
\end{center}

\subsection{The HHML dealing with an astronomy data set}
\begin{figure}[htb]\centering
\includegraphics[width=\columnwidth]{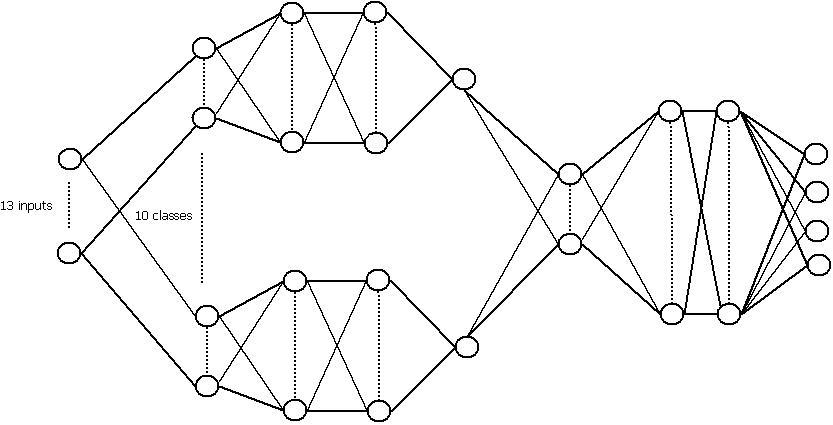}
\caption{\small Graphical scheme of the S2AN2 for the astronomical data set. In the first hyper layer there are 10 UBPs according to the 10 classes, where results of these units fed into an UBP in the second hyper layer.} \label{fig3}
\end{figure}

The astronomy dataset is a set of information extracted from the results of processing stars light curves. These data sets include all the information that can be extracted from the light curves, where these information or features capture all classes of stars and previously \cite{c17} have been employed for distinguishing between different types of stars (SMC) \cite{c15}. Since magnitude of data sets is massive and is growing every second, determining significant features from the extracted features becomes a vital problem to solve. In this experiment, the HHML method is trained on 10000 objects (stars) from the OGLE mission \cite{c16}, where 13 features represent them inside of 10 classes. Based on this information, the S2AN2 structure consists of 13 input nodes for the 13 features that are connected to 10 UBP in the first hyper layer. From the first hyper layer 10 outputs fed into another UBP in the second hyper layer. The UBP in the second layer has 4 output nodes subject to 4-digit representation of the 10 class ID's. The topology of the S2AN2 is shown in Figure\ref{fig3} and details of the UBP's are tabulated in the Table \ref{tab1}. 

\begin{table}[htb]\centering
\caption{\small Structure of the S2AN2 for dealing with the astronomical data set.}\label{tab1}
\begin{tabular}{|c|c|c|c|c|}
\hline & $\sharp$input& $\sharp$ hidden& $\sharp$ hidden& $\sharp$output\\
& nodes & nodes & nodes & nodes\\
\hline Hyper-Layer(1) & 13 & 15 & 2& 1\\
\hline Hyper-Layer(2) & 10 & 15& 20 & 4\\
\hline
\end{tabular}
\end{table} 

\begin{figure}[htb]\centering
\includegraphics[width=\columnwidth]{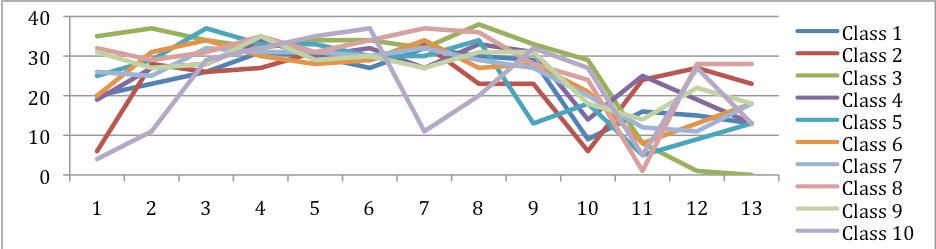}
\caption{\small In this figure, amount of corresponding weighs of every feature for each class is indicated. The horizontal axis shows features and vertical axis shows amount of importance of corresponding feature for each colored class. These values are representative of importance of a feature for distinguishing a class.From this figure, the last three features have lower importance rank rather than others.} \label{fig4}
\end{figure}

Observing amount of collaboration of features based on weights of edges in the trained network (Figure\ref{fig4} and Table \ref{tab2}) show that with the first 8 features we could capture all the classes. The refined network has applied on another test SMC data set with 40,000 stars and the results of the classification demonstrated precision of the reduced set of features. The classification processes on the test data set are performed and resulted precisely which is presented in the Table \ref{tab3}.

\begin{table}[htb]\small
\caption{\small Normalization of the values from figure 4 into [0 1] is listed in this table. Using a cutoff value like 0.5 will remove last four features. This cutoff value is examined and demonstrated as a precise cutoff value.}\label{tab2}
\begin{tabular}{|c|c|c|c|c|c|c|c|c|c|c|c|c|c|}
\hline &F.1&F.2&F.3&F.4&F.5&F.6&F.7\\
\hline Norm&5.74&7.03&8.03&8.42&8.2&8.2&7.76\\
\hline &F.8&F.9&F.10&F.11&F.12&F.13&\\
\hline Norm & 7.92&7.24&4.89&3.11&4.53&4.13&\\
\hline
\end{tabular}
\end{table} 

\begin{table}[htb]\footnotesize
\caption {\small Accuracy of the classification process (via the S2AN2) on a test dataset using first eight features. The weakness on results of classes 9 and 10 is because there exist a few samples of them in the training data set (13 and 50 samples respectively).}\label{tab3}
\begin{tabular}{|c|c|c|c|c|c|c|c|c|c|c|}
\hline &Class1&Class2&Class3&Class4&Class5\\
\hline Accuracy of & & & & &\\
 the refined & 1.15 & 1.13 & 1.34 & 1.2& 1.4  \\ \cline{2-6}
 over original DB &Class6&Class7&Class8&Class9&Class10\\ \cline{2-6}
 & 1.27 & 1.21 & 1.18 & 1& 0.98\\
\hline
\end{tabular}
\end{table} 
\subsection{The HHML performance using a plant biology data set}
\begin{figure}[htb]\centering
\includegraphics[width=\columnwidth, scale=.15]{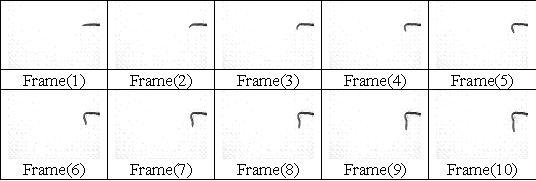}
\caption{\small Movies include 10 continuous frames. One sample movie's frames represent the movie's dynamics. As it clear from the frames, just small portion of each frame is significant and all backgrounds are as same without carrying any information. Hence, it is reasonable to expect a great dimensionality reduction in the data from this and similar redundancies. } \label{fig5}
\end{figure}

\begin{figure}[htb]\centering
\includegraphics[width=\columnwidth]{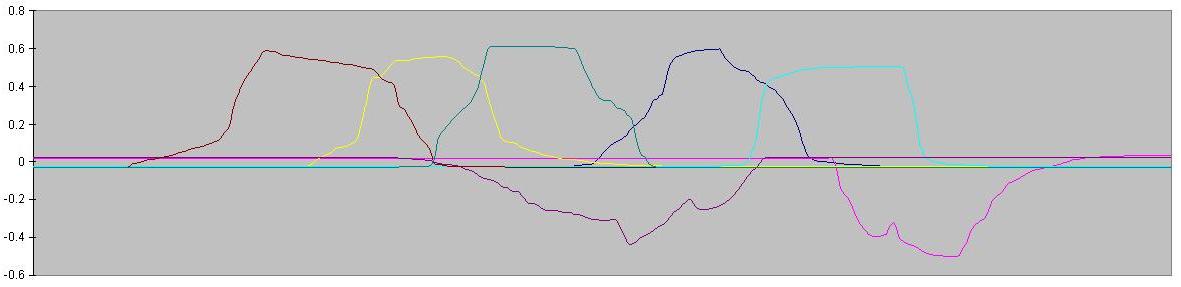}
\caption{\small Six projected frames are graphed. Projected frames have two similar areas in the beginning and at the end of the frames. These peaks are representatives of significant part of the frames, which are the root against the background.} \label{fig6}
\end{figure}

\begin{figure}[htb]\centering
\includegraphics[width=\columnwidth]{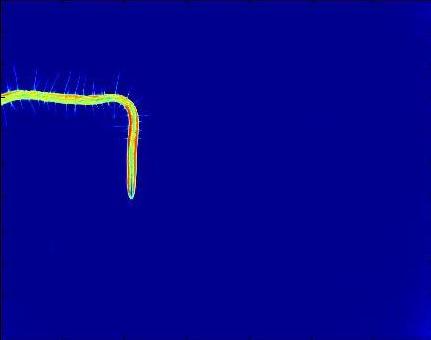}
\caption{\small All weights associated to every pixel have summed up to reach an importance matrix for pixels based on corresponding edges. Projecting a frame to this matrix resulted the left figures, where the blue colored part came from '0' importance value.} \label{fig7}
\end{figure}

This data set includes 500 movies of growth of Arabidopsis Thaliana seedlings. We chose equal number of movies from wild type and mutant roots, where 400 of them is used for training process and evaluation was performed on the rest 100 movies. Each movie is composed of 10 frames, which are used to distinguish between the classes. [The concept of predicting genotypic modification of quantitative phenotypic traits is a well-known concept and is called Phenotype to Genotype mapping \cite{c18}]. Hence, these frames are assumed as the proposed features of objects in each class and the HHML method will determine which subset of the features is representative of all the classes. A sample movie is shown in the Figure\ref{fig5}.
Each frame is a 740x740 matrix of a root shape and a vector representation of this matrix is used for feeding into the network. So in this case, each feature instead of being a value is an array and the S2AN2 uses an extra UBP for each feature (a priori to UBP's of classes) in the first hyper-layer. In order to handle this type of features the applied topology for the S2AN2 is as follows:
\begin{itemize}
\item[1] An UBP for handling a feature array. (10 feature so 10 UBPs -UBP(1)- in the first hyper layer).
\item[2] All UBP(1) feed their outputs to two UBPs corresponding to the two classes (2 classes, so 2 other UBPs -UBP(2)- in the first hyper layer).
\item[3] Results of UBP(2) feed a UBP -UBP(3)- in the second hyper layer towards calculating class IDs.
\end{itemize}
In this structure all UBPs use a sigmoid function and structure of each UBP -according to the features information- is illustrated in Table \ref{tab4}.
\begin{table}[htb]\centering
\caption{\small The S2AN2 topology for the plant biology data set.}\label{tab4}
\begin{tabular}{|c|c|c|c|c|}
\hline & $\sharp$input& $\sharp$ hidden& $\sharp$ hidden& $\sharp$output\\
& nodes & nodes & nodes & nodes\\
\hline UBP(1) & 165549 & 5000 & 2500 &1\\
\hline UBP(2) & 10 & 15& 5 & 2\\
\hline UBP(2) & 2 & 5& 2 & 1\\
\hline
\end{tabular}
\end{table} 
The error of the second layer back propagates to update weights of edges in both hyper layers. Whereas big portion of the frames is blank, the UBP(1) is employed to determine which part of the input vector is more significant. Visualization of insignificant part can be seen in Figure\ref{fig6}; vectors of six frames are projected into a two dimensional space. In fact, in this problem the S2AN2 performs two steps of dimension reduction process, it simultaneously reduces dimension of each feature (frame) and in addition extracts significant features from the 10 proposed features. Both reduction processes rely on the HHML concept and use weights of the trained S2AN2 to refine the database.
When the training process finishes, observing the weights of edges in all UBP(1) shows which part of the input vector (the image) is meaningful so the vector dimension should be reduced accordingly. Summing up all the weights in all the UBP(1) networks and converting the achieved vector to a matrix schema shows the significant part of the images. This matrix shows level of importance of the image pixels and observing this matrix shows that the frames can be reduced to 417x397 matrices, where all the significant pixels in all frames are included. Mapping a frame into this matrix is represented in Figure \ref{fig7}. 
Moreover, our experiments showed that from the 10 proposed features we just need 4 frames (5, 6, 9,10) to classify the movies. We reconstruct the S2AN2 for the reduced features and examined the network on the test part of the data set (50 movies for each class). Calculating Mean and Variance values of the edges weights gives a measure of the distribution of weights. Our cutoff value is equal to Mean-0.3*Variance, where lower weights changed to zero (associated nodes and edges removed from the S2AN2). Table \ref{tab5} shows accuracy of the network on the test data set and Table \ref{tab6} shows the ANN's resource usages when the ANN was performed on original and refined databases. The resource usages and accuracy comparisons demonstrate our algorithms practicality for analyzing biological databases, where on this feature space, fast algorithms can be run using a small amount of RAM.
\begin{table}[htb]\centering
\caption{\small Accuracy of the algorithm on the refined over original data set.}\label{tab5}
\begin{tabular}{|c|c|c|}
\hline Accuracy & 100 Movies (50 Training &100 Movies (50 Training \\
refined over & 50 Testing) from&50 Testing) from\\
original DB   & the training DB&the testing DB\\
\hline WildType & 1.03 & 1.18\\
\hline Mutated & 1.4 & 1.52\\
\hline
\end{tabular}
\end{table} 
\begin{table}[htb]\centering
\caption{\small Radiant result of comparing the ANN time and memory usage.}\label{tab6}
\begin{tabular}{|c|c|c|}
\hline Resources & Time(s) &Memory(G) \\
\hline Original DB & 763 & 8.6\\
\hline Refined DB & 286 & 3.4\\
\hline
\end{tabular}
\end{table} 
\section{Conclusion}
In this research, a new viewpoint towards tackling complex feature extraction problems is proposed. The inspiration comes from the historical advances in understanding and modeling intelligent behavior, which includes feature extraction as a cornerstone. This preliminary progress report has focused on the results of computational and algorithmic design for modeling and realization the conceptual framework. The theoretical considerations are relegated to an upcoming companion article. The computations above are based on the prevalent BP-ANN training architectures and their well-established learning abilities. The training process is modeled after the concept of a 'pipeline' where inputs are processed with the additional provision of determining inputs' role in calculating the results of the output nodes. We evaluated our method using two different databases of movies from biological systems and astronomical observations. The assessment poses that the architecture extracts the biologically significant parts of the frames, and provided a novel method for 'dimensionality and size reduction' by orders of magnitude for movies in the plant biology data set. Manipulating an entirely different pattern recognition task, an astronomy data set was studied using our model. The significant astronomy data features were also extracted. The latter outcomes are along the lines of the scientific objectives to provide helpful software tools for future cosmology missions. We are confident that our method is applicable to other domains and classes of similar problems. Our work in progress includes development of an improved version that could be applied to solve problems using the ultimate power of the parallel processing platforms. Also, the project is continuing with positive progress to solve practical applications for dimensionality reduction on-demand for astronomy data sets as they are becoming available by the GAIA \cite{c19} mission in the European Space Agency \cite{c20}.
\section*{Acknowledgement}
The authors wish to thank Professor Luis S. Baro for his valuable helps on understanding the astronomical problem and providing the data sets. The authors also wish to thank Nathan Miller and Tessa Durham (Department of Botany, University of Wisconsin-Madison) for samples of the root growth movies.

\end{document}